# PENELOPIE: Enabling Open Information Extraction for the Greek Language through Machine Translation

**Dimitris Papadopoulos**♣♦, **Nikolaos Papadakis**♦ and **Nikolaos Matsatsinis**♣
♣Technical University of Crete, Greece
♦Hellenic Army Academy, Greece
{dpapadopoulos6, nmatsatsinis}@isc.tuc.gr
npapadakis@sse.gr

## Abstract

In this work, we present a methodology that aims at bridging the gap between high and low-resource languages in the context of Open Information Extraction, showcasing it on the Greek language. The goals of this paper are twofold: First, we build Neural Machine Translation (NMT) models for English-to-Greek and Greek-to-English based on the Transformer architecture. Second, we leverage these NMT models to produce English translations of Greek text as input for our NLP pipeline, to which we apply a series of pre-processing and triple extraction tasks. Finally, we back-translate the extracted triples to Greek. We conduct an evaluation of both our NMT and OIE methods on benchmark datasets and demonstrate that our approach outperforms the current state-of-the-art for the Greek natural language.

## 1 Introduction

Open Information Extraction (OIE) techniques generally shine in high-resource languages (e.g. English, German) for which either linguistic principles leading to triple extraction have been identified or large annotated corpora and pre-trained language models can be used. For low-resource languages like Modern Greek however, there is a relative sparsity of raw textual resources and annotated corpora that could lead to the development of similar systems. On the bright side, the need for multilingual resources (e.g. movie subtitles, applications, web content) has fueled several projects of compiling parallel corpora (i.e. collections of texts translated into one or more other languages than the original) over the last years. In this work, we propose a methodology that aims at enabling OIE for low-resource languages, focusing on the Greek OIE use case. To achieve this, we rely on Neural Machine Translation (NMT) as an intermediate step to translate the texts to English, in order to exploit the plethora of methods that exist for transforming English text to its structured representation.

We present PENELOPIE (Parallel EN-EL Open Information Extraction), a pipeline for information extraction from Greek corpora. An overview of our methodology is given in Figure 1. The code and related resources can be found in https://github.com/lighteternal/PENELOPIE.

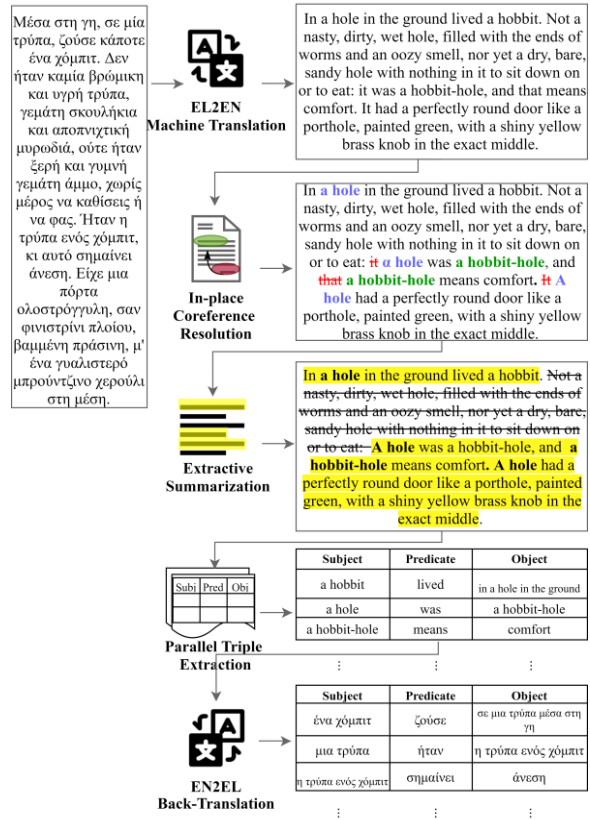

Figure 1: Steps of the PENELOPIE pipeline

Our study has the following objectives:

1. to release a series of Transformer-based NMT models for English-to-Greek (EN2EL) and Greek-to-English (EL2EN) translation, trained on a consolidated parallel corpus and compare the translation results to the current-state-of-the-art,
2. to leverage the aforementioned NMT models for the translation of Greek texts to their English counterpart and feed them to our English-based NLP pipeline. This pipeline incorporates a series of pre-processing tasks including in-place coreference resolution and extractive summarization, as well as an OIE system comprising of three extractors based on different approaches for more robust results. The extracted triples are finally back-translated to Greek and their quality is evaluated to facilitate comparison with other methods.

## 2 Background

In this section, we provide background information on neural machine translation and open information extraction approaches.

### 2.1 Neural Machine Translation

NMT aims at modelling a direct mapping between source and target languages with deep neural networks. It has become the dominant paradigm of machine translation, achieving promising results in recent years which are usually surpassing those of traditional Statistical Machine Translation (SMT) approaches, given enough training data (Stahlberg, 2020). The invention of novel encoder-decoder architectures, from recurrent (Sutskever et al., 2014; Bahdanau et al., 2015) and convolutional neural networks (Kalchbrenner et al., 2014; Gehring et al., 2017) to self-attention (Transformer) mechanisms (Vaswani et al., 2017; So et al., 2019) has significantly pushed ahead the state-of-the-art, in terms of quality and efficiency, especially for morphologically rich languages.

Another parallel line of research towards improving translation quality is to devise effective token encoding methods that can handle out-of-vocabulary (OOV) words, targeting the lack of 1-to-1 correspondence between source and target languages, due to differences in their morphological structure. Sennrich et al. (2016) utilized variants of byte-pair encoding (BPE) methods for word segmentation to enable the representation of rare and unseen words as a sequence of subword units, showing that NMT methods are capable of open-vocabulary translation. The latest advances in the field also include pretraining cross language models on multilingual data (Conneau and Lample, 2019) or exploiting monolingual corpora for semi-supervised learning through back-translation (Sennrich et al., 2016a). It appears that not much effort has been targeted towards the Greek language with the notable exception of the Helsinki NLP group which has released EN2EL and EL2EN translation models evaluated on the Tatoeba dataset (Tiedemann and Thottingal, 2020).

### 2.2 Information Extraction

Open information extraction (OIE) systems aim at distilling structured representations of information from natural language text, usually in the form of {subject, predicate, object} triples or n-ary propositions. Since OIE follows a relation-independent extraction paradigm, it can play a key role in many NLP applications including natural understanding and knowledge base construction, by extracting phrases that indicate semantic relationships between entities. In order to extract triples, most approaches try to identify linguistic extraction patterns, either hand-crafted or automatically learned from the data. An abundance of such systems exists, relying on concepts ranging from rule-based paradigms that focus on the grammatical and syntactic properties of the language (Fader et al., 2011; Del Corro and Gemulla, 2013), to supervised learning-based ones that leverage annotated data sources to train classifiers, with more recent implementations making use of language models (Kolluru et al., 2020; Ro et al., 2020). Despite the existence of so many approaches however, the majority of them just focuses on evaluating the efficiency of different triple extraction tools on raw data, without incorporating any preprocessing strategies to limit the number of potentially uninformative triples (Niklaus et al., 2018). Some more recent methods go beyond the triple extraction task by encompassing more thorough preprocessing and postprocessing strategies, including discourse analysis, coreference resolution or summarization to improve the quality of the extracted triples (Kertkeidkachorn and Ichise, 2017; Papadopoulos et al., 2020). There is currently no OIE system for the Greek language, although latest approaches that leverage pretrained language models allow for multilingual extractions through zero-shot learning (Ro et al., 2020).

## 3 Methodology

The NMT architecture used in this paper for English-to-Greek (EN2EL) translation and Greek-to-English (EL2EN) back-translation is a variant of the Transformer model (Vaswani et al., 2017), driven by the fact that self-attentional networks tend to perform distinctly better than other architectures on translation tasks (Tang et al., 2020). Both the encoder and the decoder are composed of stacked, multi-head, self-attention and fully-connected layers. One key difference between the two implementations is that ours includes a fully connected feed-forward network with an inner-layer dimensionality of $d_{ff} = 1024$, as opposed to the original one that uses a hidden layer with $d_{ff} = 2048$, in an effort to reduce computational cost, as our training testbed had limited memory capabilities. With regard to vocabulary construction, we relied on subword units extracted with BPE, experimenting with two different configurations of merge operations

Our approach for efficient open information extraction on translated texts combines a series of distinct modules for in-place coreference resolution, extractive summarization and parallel triple extraction with the following specifications:

**Coreference Resolution:** We rely on a variant of the pretrained end-to-end coreference resolution model from Lee et al. (2017) using Span-BERT embeddings (Joshi et al., 2020), trained on the OntoNotes 5.0 dataset. Each translated sequence is pre-processed by the in-place coreference resolution component, where all noun phrases (mentions) referring to the same entity are substituted with that entity.

**Summarization:** Extractive text summarization is used on the coreference-resolved text to reduce the original documents' length by omitting peripheral information while highlighting key features that are appropriate for triple extraction. We use the transformer-based implementation from Miller (2019) where all sentences are embedded into the multi-dimensional space using BERT embeddings. K-means clustering is then used on the sentence representations to identify those closest to the cluster's centroids for summary selection.

**Parallel Triple Extraction:** Here we combine three popular OIE systems, relying both on rule-based (handcrafted extraction heuristics and clauses) and learning-based (semantic role labelling and sequence BIO tagging) systems, relying on the complementarity between the different approaches to ensure maximum recall.

We provide additional information regarding the technical implementation of the described information extraction pipeline in the following section.

## 4 Experimental Setup

### 4.1 NMT Setup

**Dataset:** We exploited most of the EN-EL resources available in the OPUS repository (Tiedemann, 2012), with main ones being the ParaCrawl, OpenSubtitles, EUBookshop, DGT and Europarl datasets. We combined these with the available parallel corpora of CCMatrix (Schwenk et al., 2019) mined in the textual content of Wikipedia, to create a dataset comprising 50451352 sentences (~6.3GB).

**Preprocessing:** We applied a cleaning script on the corpus that discarded any segment with a word exceeding 1000 characters, leading to a corpus of 36251157 sentences. We tokenized these using the Moses[1] tokenizer and split the dataset so that 1 every 23 sentences were assigned to the validation set and the rest to the training set. For the construction of the model dictionaries, we worked towards the creation of two different preprocessing setups leading in two training configurations:

a. For the first setup, we lower-cased all tokens in the train and test set, in an effort to reduce the dictionary size without losing translation quality. We then applied BPE segmentation[2] with an encoding size of 10000 to speed up training and inference. This resulted in dictionaries of 12892 and 9932 tokens for Greek and English accordingly.
b. For the second setup, we applied BPE segmentation directly to the mixed-case text with an encoding size of 20000,

---

[1] https://github.com/moses-smt/mosesdecoder

[2] https://github.com/rsennrich/subword-nmt

resulting in dictionaries of size 23220 and 15284 for Greek and English respectively.

**NMT Model Settings and Training:** We utilized Fairseq (Ott et al., 2019), a popular sequence-to-sequence toolkit maintained by Facebook AI Research to train our models with data from both setups and ran our experiments on a machine with a single NVIDIA GeForce RTX-2080 SUPER (8GB of VRAM). We implemented a shallower variant of the Transformer architecture with 4 attention heads, 6 encoder and 6 decoder layers, both with an embedding size of 512 and a feed-forward hidden layer dimension of 1024. During training, regularization was done with a dropout of 0.3 and label smoothing of 0.1. We used the Adam optimizer (Kingma and Ba, 2015) with 4000 warm-up steps and a maximum learning rate of 0.0005. The model was trained for 5 epochs and the best checkpoint was selected based on the perplexity of the validation set. We used mixed precision during training (Narang et al., 2018), using FP16 precision to address our hardware limitations by reducing the memory consumption and time spent in memory. The produced models (4 in total) are as follows: i. a lower-case EL2EN and a lower-case EN2EL model from the first setup based on shorter dictionaries, ii. a mixed-case EL2EN and EN2EL model from the second setup on larger dictionaries.

### 4.2 Information Extraction Setup

**Coreference Resolution Framework:** Each EL2EN translated sequence was processed by the pretrained neural model from AllenNLP which relies on Lee et al. (2017) but has the original GloVe embeddings substituted with Span-BERT embeddings. This approach considers all possible spans in a document as potential mentions and learns distributions over possible antecedents for each span. Its ability to solve challenging pronoun disambiguation problems facilitated the creation of more informative triples.

**Summarization Framework:** In order to reduce the size of the ingested text, we relied on the pretrained extractive summarizer from Miller (2019) made available by HuggingFace, that utilizes the BERT model for text embeddings and k-means clustering to identify sentences close to the centroid for summary selection.

**Triple Extraction Engines:** We integrated 3 OIE engines based on different extraction strategies: a. Open IE 5.1 from UW and IIT Delhi which is based on the combination of four different rule-based and learning-based OIE tools, b. ClausIE from MPI that follows a clause-based approach, and c. AllenNLP OIE that formulates the triple extraction problem as a sequence BIO tagging problem and applies a bi-LSTM transducer to produce OIE tuples. We further employed a deduplication process to keep only the unique triples and eliminate all redundant extractions. Since the goal of our work was to provide triples in the Greek language and the produced triples were in English, we used our EN2EL NMT model to translate them back to Greek.

## 5 Results and Discussion

We provide results both for the EL-EN NMT tasks and for the OIE task on Greek corpora, since the former can be evaluated independently.

### 5.1 NMT performance

Table 1 shows the evaluation of our models (lower-case and mixed-case) on the Tatoeba[3] and XNLI[4] test sets.

| *Evaluation on Tatoeba test set (EN-EL)* | | |
|---|---|---|
| **Model** | **BLEU** | **chrF** |
| Helsinki-2019-12-04-*EN2EL* | 52.7 | 0.721 |
| Helsinki-2019-12-18-*EN2EL* | 56.4 | **0.745** |
| **OURS**-lower-case-*EN2EL* | **77.3** | 0.739 |
| **OURS**-mixed-case-*EN2EL* | 76.9 | 0.733 |
| Helsinki-2019-12-04-*EL2EN* | 69.4 | 0.801 |
| **OURS**-lower-case-*EL2EN* | **79.9** | **0.802** |
| **OURS**-mixed-case-*EL2EN* | 79.3 | 0.795 |
| *Evaluation on XNLI test set (EN-EL)* | | |
| **Model** | **BLEU** | **chrF** |
| **OURS**-lower-case-*EN2EL* | 66.1 | 0.606 |
| **OURS**-mixed-case-*EN2EL* | 65.4 | 0.624 |
| **OURS**-lower-case-*EL2EN* | 67.4 | 0.633 |
| **OURS**-mixed-case-*EL2EN* | 66.2 | 0.623 |

Table 1: EN2EL and EL2EN NMT evaluation results & comparison with other models.

For both EN-EL and EL-EN directions we compare with the current state-of-the-art models produced by the Helsinki NLP group, evaluated on the Tatoeba dataset (Tiedemann and Thottingal, 2020). Another relevant implementation from the Facebook AI team provides results of their XLM-R model on the XNLI dataset (Ruder et al., 2019); however -given the different scope of that paper- results are presented in terms of cross-lingual classification accuracy and not in terms of NMT translation quality (e.g. BLEU), hindering direct comparisons. Nevertheless, we also provide BLEU and chrF scores on the parallel EN-EL corpus of the XNLI dataset hoping that it will facilitate comparisons with future models.

---

[3] https://tatoeba.org/

[4] https://github.com/facebookresearch/XNLI

The results on the Tatoeba test set showcase a significant performance gain of our models in terms of BLEU (+10.9 BLEU for EN2EL and +10.5 BLEU for EL2EN translations) over the Helsinki ones, while all models have very close chrF scores. The apparent difference in performance gains between the two different metrics can ascribed to the idiosyncratic morphological and syntactic properties of the Greek language (accent, inflation, declension etc.) that may result in the produced translations being slightly different from the original sequences. Since chrF incorporates character matches while BLEU does not, it is possible to produce translations that achieve low BLEU but acceptable chrF scores. Therefore, given that BLEU is an n-gram-based metric and chrF is a character-based one, we consider the good results on both metrics as a positive characteristic towards producing quality estimates that are as close as possible to human judgements. The results also seem promising on the more challenging XNLI test set, although a direct comparison with other models would have been more useful. While the lower-case models seem to perform slightly better on every test, the richer vocabulary and correct casing of the mixed-case ones compensates for the slightly worse metrics scores. It should be noted that in order to ensure a fair comparison, the mixed-cased models were evaluated on the original reference translations, while the lower-case models were evaluated on a lower-case version of the same translations. Another aspect that adds to the reason why lower-case NMT models were able to showcase slightly better scores is that the former reduce the expansion of the vocabulary by neglecting some morphology information, while mixed-case models will increase the vocabulary to keep the original morphological form and as a result may lose connections with the lowercase forms of some words. Finally, while our models were trained using the Fairseq framework, we also ported them to HuggingFace Transformers format and made them publicly available[5].

## 5.2 OIE performance

We evaluate the performance of PENELOPIE using the CaRB benchmark which is widely used for the comparison of OIE systems (Bhardwaj et al., 2020). Given the lack of a gold standard of Greek annotated triples, we created a translated version of the original CaRB test set for our experiments, consisting of 2715 sentences and their extracted semantic triples. The test set was automatically translated using our EN2EL mixed case model. We compare our extraction results with Multi2OIE from Ro et al. (2020), an OIE engine with state-of-the-art performance on English corpora. Multi2OIE relies on the pretrained multilingual BERT model and can perform multilingual extractions through zero-shot learning (it is trained on English data); thus it can be leveraged to produce results on the Greek CaRB test set. For PENELOPIE, results are only provided using the mixed-case NMT model (similar results to the lower-case one). It should be noted that the summarization module was not utilized during the benchmark, as the gold dataset consisted of single sentences. This is a general shortcoming in the assessment of OIE systems that leverage preprocessing features (such as summarization or coreference resolution); the gold triples and the metrics involved in the evaluation process favour exact matches of the processed sentences, rather than focusing on the usability of the extracted results. As a result, some of the gold triples in benchmark datasets -although valid- may have low contextual value. The scores are presented in terms of precision, recall and F1-score in Table 2:

| Model | Prec. | Rec. | F1 |
|---|---|---|---|
| Multi2OIE | 0.200 | 0.084 | 0.118 |
| **PENELOPIE** | **0.231** | **0.284** | **0.255** |

Table 2: PENELOPIE evaluation results on the translated CaRB testset & comparison with Multi2OIE.

---
[5] https://huggingface.co/lighteternal

Our pipeline outperforms the state-of-the-art Multi2OIE on the Greek OIE task, on all metrics. The most remarkable difference in performance is shown in terms of recall, which can be partially attributed to the fact that PENELOPIE leverages a number of different extraction tools leading to a recall-oriented approach. In addition, given that all triples are individually back-translated to Greek, it is not guaranteed that the translation output of each element will match the span of the derived sentence, especially in languages with rich morphology (e.g. conjugation, declension). This justifies the relatively low scores of PENELOPIE compared to English OIE systems, whose F1-scores may exceed 0.50 for state-of-the-art approaches (although a direct comparison between different languages is not straightforward). To this end, a source-target word alignment approach inspired by the work of Garg et al. (2020) was explored, but current implementations seem to have difficulties in aligning tokens with accents[6] (e.g. Greek ones).

## 6 Conclusions and Future Work

We have presented the use of NMT models integrated in an OIE pipeline to achieve triple extraction for low-resource languages, showcasing our approach on the Greek language. To this end, we trained 4 models (2 EN2EL and 2 EL2EN) that outperform the state-of-the-art by a significant margin (>10 BLEU) and made them publicly available. We leveraged these along with a set of preprocessing and triple extraction tools to construct the PENELOPIE pipeline aiming at information extraction from Greek texts. We demonstrated the efficiency of our methodology via a benchmark framework and obtained significantly better results (+116% in F1-score) compared to the best multilingual OIE system currently available.

For future work, we will focus more on word-level alignment to improve the quality of our extractions. We would also like to explore transfer learning approaches to create an end-to-end OIE system for Greek without relying on annotated datasets.

## Acknowledgments

The research work of D.P. was supported by the Hellenic Foundation for Research and Innovation (HFRI) under the HFRI PhD Fellowship grant (Fellowship Number: 50, 2nd call).

---

[6] https://github.com/lilt/alignment-scripts